\documentclass[10pt,twocolumn,letterpaper]{article}

\usepackage{cvpr}
\usepackage{times}
\usepackage{epsfig}
\usepackage{graphicx}
\usepackage{amsmath}
\usepackage{amssymb}
\usepackage{graphicx}
\usepackage{algorithm}
\usepackage{algorithmic}
\usepackage{subfigure}

\usepackage[breaklinks=true,bookmarks=false]{hyperref}

\cvprfinalcopy 

\def\cvprPaperID{****} 

\begin{document}

\title{IRLAS: Inverse Reinforcement Learning for Architecture Search}

\author{Minghao Guo\\
The Chinese University of Hong Kong\\
{\tt\small gm09@ie.cuhk.edu.hk}
\and
Zhao Zhong\\
NLPR, CASIA\\
University of Chinese Academy of Sciences\\
{\tt\small zhao.zhong@nlpr.ia.ac.cn}
\and
Wei Wu\\
SenseTime Research\\
{\tt\small wuwei@sensetime.com}
\and
Dahua Lin\\
The Chinese University of Hong Kong\\
{\tt\small dhlin@ie.cuhk.edu.hk}
\and
Junjie Yan\\
SenseTime Research\\
{\tt\small yanjunjie@sensetime.com}
}

\maketitle
\begin{abstract}
In this paper, we propose an inverse reinforcement learning method for architecture search (IRLAS), which trains an agent to learn to search network structures that are topologically inspired by human-designed network. Most existing architecture search approaches totally neglect the topological characteristics of architectures, which results in complicated architecture with a high inference latency. Motivated by the fact that human-designed networks are elegant in topology with a fast inference speed, we propose a mirror stimuli function inspired by biological cognition theory to extract the abstract topological knowledge of an expert human-design network (ResNet). To avoid raising a too strong prior over the search space, we introduce inverse reinforcement learning to train the mirror stimuli function and exploit it as a heuristic guidance for architecture search, easily generalized to different architecture search algorithms. On CIFAR-10, the best architecture searched by our proposed IRLAS achieves 2.60\% error rate. For ImageNet \emph{mobile} setting, our model achieves a state-of-the-art top-1 accuracy 75.28\%, while being 2$\sim$4$\times$ faster than most auto-generated architectures. A fast version of this model achieves 10\% faster than MobileNetV2, while maintaining a higher accuracy.
\end{abstract}

\section{Introduction}

The past several years have witnessed the remarkable success of convolutional neural networks in computer vision applications. Thanks to the advances in network architectures, \eg~ResNet~\cite{he2016deep}, Inception~\cite{szegedy2016rethinking} and DenseNet~\cite{huang2017densely}, the performances on a number of key tasks, such as image classification, object detection, and semantic segmentation, have been taken to an amazing level. 
However, every step along the way of network design improvement requires extensive efforts from experienced experts and takes a long period of time. This already constitutes a significant obstacle to further progress.

Naturally, automatically finding suitable network architectures for a given task becomes an alternative option and is gaining ground in recent years. Along this direction, a number of network search methods have been developed, including evolution \cite{saxena2016convolutional,xie2017genetic}, surrogate model based search \cite{liu2017progressive,perez2018efficient}, and reinforcement learning \cite{zoph2016neural,zoph2017learning,zhong2018practical,baker2016designing}. Whereas these methods have shown promising results and found new architectures that surpass those crafted by experts, they are still subject to a serious limitation -- auto-generated networks usually have a rather high inference latency, making them difficult to be deployed on practical system with limited computational capabilities. An important cause to this issue is that auto-generated structures are often excessively complicated, which, as observed in~\cite{ma2018shufflenet}, tends to adversely influence the run-time efficiency.
While there have been attempts \cite{tan2018mnasnet} to incorporate latency information to guide the search, the problem has not been effectively solved -- the search algorithms themselves still follow a pre-defined way for network motif construction, \eg~recursively expanding a tree structure as in NASNet~\cite{zoph2017learning}, without enforcing any explicit guidance to the network topology.

\begin{figure}[tb]
  \begin{center}
    \includegraphics[width=1.0\linewidth]{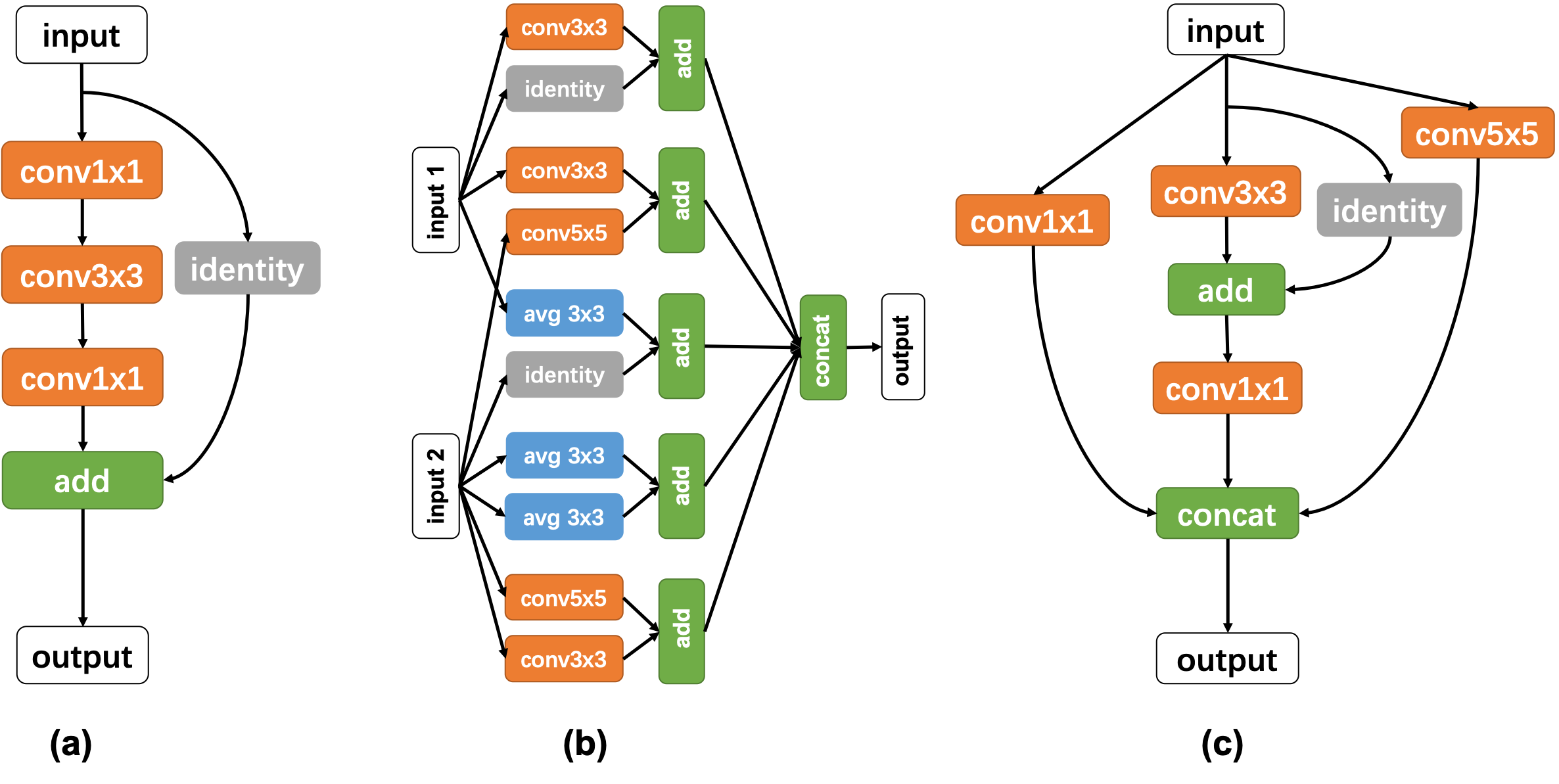}
  \end{center}
  \caption{\small{Topologies of different architectures. Human-designed architectures have a more simple and elegant topology than existing auto-generated architectures. Our IRLAS aims to search topologically elegant architectures guided by human-designed networks. (a) ResNeXt \cite{xie2017aggregated}; (b) NASNet \cite{zoph2017learning}; (c) Best performed architecture found by our IRLAS. }}
  \label{fig-motivation}
  \vspace{-16pt}
\end{figure}

\begin{figure*}[tb]
  \begin{center}
    \includegraphics[width=1.0\linewidth]{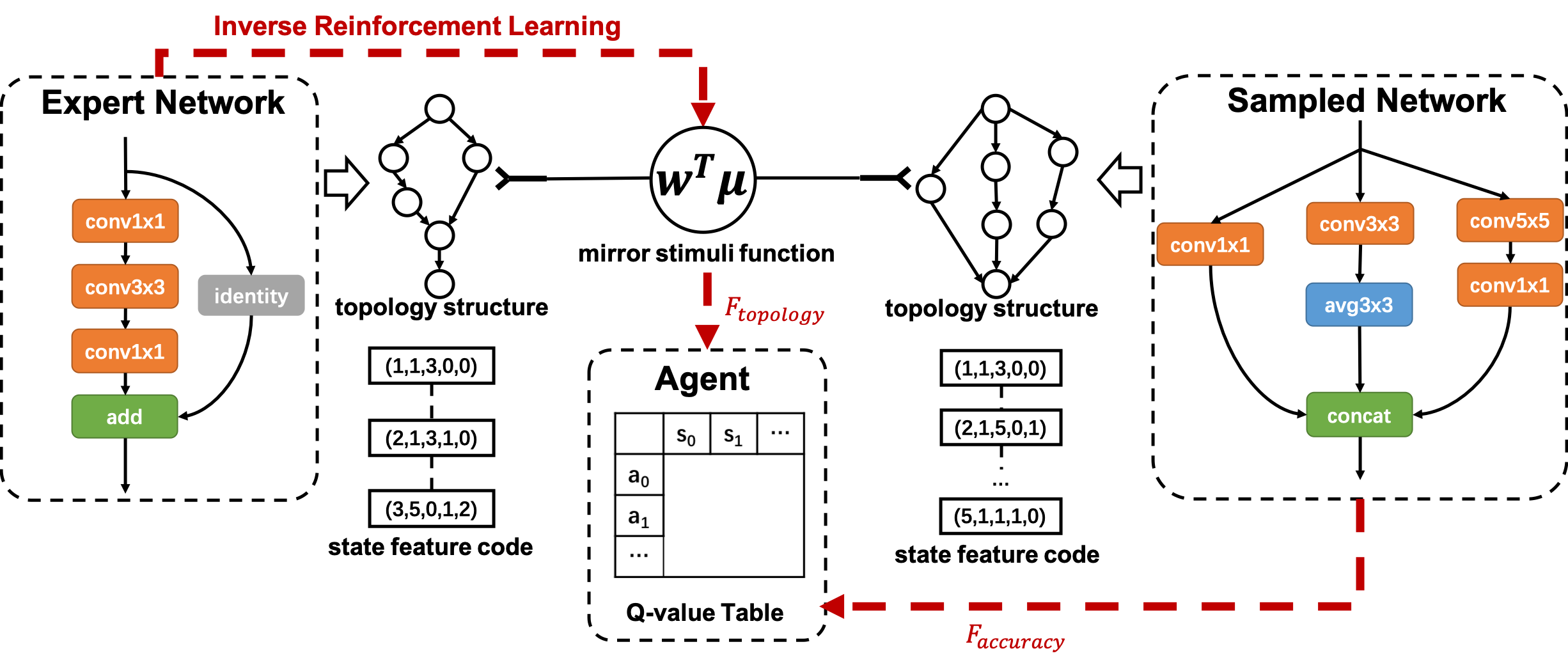}
  \end{center}
  \caption{\small{The pipeline of our IRLAS. We propose a mirror stimuli function to extract the abstract representation for topological characteristic of the expert. Topology structures of networks are converted to state feature code as the input of mirror stimuli function. During the agent's searching process, the mirror stimuli function is utilized as a heuristic guidance to generate desirable human-designed-like networks. Inverse reinforcement learning is utilized to train the mirror stimuli function, which helps the agent to efficiently explore the large search space without being overly restricted.}}
  \label{fig-pipeline}
  \vspace{-12pt}
\end{figure*}

In this work, we aim to explore a new approach that explicitly takes the topological structure into account.
Our efforts are motivated by the observation that human-designed networks are usually topologically simple, 
as shown in Figure \ref{fig-motivation}, especially when compared to auto-generated ones, and often stride a better balance between accuracy and efficiency. These designs are often grounded on the rich experiences obtained through many years of joint efforts by the community, which are valuable resources and deserve to be leveraged during the searching process.
 
Specifically, we propose an inverse reinforcement learning method for architecture search (IRLAS). 
At the heart of this method is a mirror stimuli function learned by inverse reinforcement learning. This function is expected to reward those architectures that are topologically similar to the networks designed by experts. During the searching process, an agent resorts to this function to provide structural guidance, so as to generate networks with desirable architectures, similar to those crafted by experts.

This method has two benefits:
(1) While the search receives guidance from the mirror stimuli function, it is not restricted. The agent is allowed to explore instead of just copying the experts.
(2) The mirror stimuli function is generic and is orthogonal to the design of search space and strategy. Hence, it can be readily generalized to different search settings. 
On both CIFAR-10 \cite{krizhevsky2014cifar} and ImageNet \cite{deng2009imagenet}, IRLAS is able to find new architectures that yield high accuracies while maintaining low inference latency.

Our contributions are summarized as follows:
\textbf{1)} We propose a mirror stimuli function that can provide topological guidance to architecture search, based on the knowledge learned from the expert-designed networks. This function can be easily generalized to different architecture search algorithms.
\textbf{2)} We introduce inverse reinforcement learning algorithm to train the mirror stimuli function, which helps the agent to efficiently explore the large search space without being overly restricted.
\textbf{3)} The network searched by our IRLAS is topologically similar to the given expert network and shows competitive accuracy and high inference speed, compared to both state-of-the-art human-designed and auto-searched networks. On CIFAR-10, the best architecture searched by our proposed IRLAS achieves 2.60\% error rate. For ImageNet mobile setting, our model achieves a state-of-the-art top-1 accuracy 75.28\%, while being 2$\sim$4$\times$ faster than most auto-generated architectures. A fast version of this model achieves 10\% faster than MobileNetV2, while maintaining a higher accuracy.

\section{Related Work}

\subsection{Neural Architecture Search}

Neural architecture search focuses on automatically searching effective neural topologies in a given architecture space. Existing architecture search methods can be mainly classified into three categories: evolutionary, surrogate model based search and reinforcement learning. Evolutionary methods \cite{domhan2015speeding,xie2017genetic,real2018regularized} aim to simultaneously evolve the topology of a neural network along with its weights and hyperparameters to evolve a population of networks. 
Early evolutionary approaches utilized genetic algorithms to optimize both the architecture and its weights, while recent studies used gradient-based methods and evolutionary algorithms to optimize the weights and architecture respectively.

Surrogate model based search methods \cite{liu2017progressive,brock2017smash,perez2018efficient} utilize sequential model-based optimization as a technique for parameter optimization. 
Typical methods like PNAS \cite{liu2017progressive} performed a progressive scan of the neural architecture search space, which was constrained according to the state-of-the-art of previous iterations. 
EPNAS in \cite{perez2018efficient} further increased the search efficiency by sharing weights among sampled architectures. However, these methods generate architectures greedily by picking the top K at each iteration, which may result in a sub-optimum over the search space.

Reinforcement learning (RL) methods \cite{baker2016designing,zoph2016neural,zhong2018practical,zoph2017learning,real2018regularized} formulate the generation of a neural architecture as an agent's action, whose space is identical to the architecture search space. The agent's reward is the performance of the trained architecture on unseen data. 
Differences between different RL-based approaches lie in the representation of agent's policy and how to optimize it. For example, \cite{zoph2016neural} used a recurrent neural network (RNN) to sample a sequence of string which encoded the neural architecture. Policy gradient algorithms including REINFORCE and Proximal Policy Optimization (PPO) were used to train the agent. \cite{baker2016designing} and \cite{zhong2018practical} used Q-learning to train a policy that sequentially chose a layer's type and its corresponding hyperparameters. There are some other RL-based methods that transform existing architectures incrementally to avoid generating entire networks from scratch, such as \cite{cai2018efficient}. However, these approaches could not visit the same architecture twice so that strong generalization over the architecture space was required from the policy. Instead of directly using an existing architecture as an initialization, our IRLAS aims to learn a mirror stimuli function, and utilizes it in the searching process as a heuristic guidance without any restraints for the search space.

There also exist recent efforts \cite{liu2018darts} introducing a real-valued architecture parameter, which was jointly trained with weight parameters. Different from other methods, this kind of algorithm does not involve architecture sampling during searching process. Our mirror stimuli function can also be generalized to this brunch of methods.

\subsection{Imitation Learning}
As our proposed IRLAS attempts to generate architectures that are topologically similar to human-designed networks, the learning for the agent involves imitation learning problem. Imitation Learning (IL) enables an agent to learn from demonstrations of an expert, independent of any specific knowledge in the proposed task. There exist two different areas for IL: policy imitation and inverse reinforcement learning. Policy imitation, which is also known as behavioral cloning, targets directly learning the policy mapping from perceived environment or preprocessed features to the agent's actions. For the settings of this paper, since the number of human-designed networks is limited, it is hard to obtain sufficient number of expert's state-action tuples for supervised learning. As a result, the direct policy imitation cannot be used for our purpose.

Inverse reinforcement learning (IRL) refers to the problem of deriving a reward function from observed behavior. As it is a common presupposition that reward function is a succinct, robust and transferable definition of a task, IRL provides a more effective form of IL than policy imitation. 
Early studies in IRL \cite{abbeel2004apprenticeship,ziebart2008maximum,ratliff2006maximum} assumed that the expert was trying to optimize an unknown reward function that could be expressed as a linear combination of pre-determined features.
\cite{choi2013bayesian} extended this approach to a limited set of non-linear rewards and learned to build composites of logical conjunctions for atomic features. Other flexible non-linear function approximators such as Gaussian Processes further extended the modeling capacity of IRL models \cite{levine2011nonlinear}. In this paper, we assume the reward function of the expert network as a linear parametrization of state features. Experiments show that this simple assumption is effective enough to extract the topological knowledge of the human-designed architectures.

\section{Approach}

In this section, we first present the problem formulation of architecture search. Then we propose the mirror stimuli function inspired by biological cognition and its training procedure via inverse reinforcement learning. Finally we detail the search space and the searching algorithm. The pipeline of our IRLAS is shown in Figure \ref{fig-pipeline}.

\subsection{Problem Formulation}\label{section-formulation}

Like modern CNNs, our automatic neural network process designs the topological structure of \emph{block} instead of the entire network. This block-wise design is more flexible for different datasets and tasks with powerful generalization ability. The task of the agent is to sequentially sample layers from the pool of layer candidates to form the block. Then the block structure is stacked sequentially to form the complete network. For different datasets, we manually choose different number of down-sampling operations due to different input image size and choose different repeat times of the block to meet the demand for limitation of parameters or FLOPs.

In this paper, we consider the design process of network topology as a variable-length decision sequence for the choice of operation. And this sequential process can be formulated as a Markov decision process (MDP). The \emph{policy} $\pi: \mathcal{S} \rightarrow \mathcal{A}$, where $\mathcal{S}$ is the state space and $\mathcal{A}$ is the action space, determines the agent's behavioral preference of generating architectures. The \emph{state} $s \in \mathcal{S}$ is the status of current layer. The \emph{action} $a \in \mathcal{A}$ is the decision for the subsequent layer. Thus, an architecture $m$ sampled by the agent can be determined by a state-action trajectory according to the policy $\pi$, i.e. $m=\{ {(s_t, a_t)}\}_{t=1...T}$. 

The training of the agent is to maximize the expected \emph{reward} over all possible architectures, 
\begin{eqnarray}\label{def_J}
J_\pi = \mathbb E_{\pi}[R(m)],
\end{eqnarray}
where $R(\cdot)$ is the reward function. A common definition of $R(m)$ is the validation accuracy of the corresponding network. This formulation of the reward function is based on an assumption that the evaluation for an architecture is only determined by its validation performance, while totally neglect the topology information.

\subsection{Topological Knowledge}

As the human-designed architectures are demonstrated to be effective in practice, we attempt to utilize such existing abundant topological knowledge as efficacious guidance for architecture search. However, it is a challenging problem to find an effective method to formalize the abstract topological knowledge and design an appropriate way to further exploit it in the search process. For example, shortcut connection of the block in ResNet is a quotable structure for architecture generating. Human can easily understand the topological structure simply by visualization, while the agent cannot. It remains harder for the agent to learn to search ResNet-like architectures if it even cannot understand the topology. This naturally raises two basic problems: 1) How to encode network architecture to extract the abstract topological knowledge as an available input for the agent? 2) How to utilize this knowledge to guide the agent to design desirable architectures? 

For the first problem, we need to define a feature embedding for network architectures. To encode the architecture, we carefully choose a \emph{state feature} function $\phi: \mathcal{S} \rightarrow \mathbb{R}^{k\times 1}$, which consists of: $operation \ type$, $kernel \ size$, and the $indexes$ of two predecessor of the current layer (for layer with only one predecessor, one of the indexes is set to zero). Despite the simplicity, this state feature function provides a complete characterization of the network architecture, including the information about the computation carried out by individual layers as well as how the layers are connected.

We further exploit \emph{feature count} to unify the information of each state feature to get the feature embedding for the whole architecture. Given an architecture's sequential trajectory $m = \{ {(s_t, a_t)}\}_{t=1...T}$, the feature count is defined as:
\begin{eqnarray}\label{eq:mu}
\mu = \sum_{t=1}^{T} \gamma^{t} \phi(s_t),
\end{eqnarray}
where $\gamma$ denotes a discounted scalar. Thus, the sequential order is also included by the discounted $\gamma$ over layer index. The feature count is utilized as an appropriate encoding for the topological knowledge of a given network. 

As for the other question of how the agent uses the topological knowledge as a guidance, this encompasses the classical exploration-exploitation trade-off. We attempt the agent to search architectures that are topologically similar to the expert network, while efficiently explore the architecture search space. This requires the searching algorithm exhibiting no preferences on a specific architecture as we do not aim the agent to reproduce human-designed networks. Direct policy imitating between the feature counts of sampled architecture and expert network will raise a strong prior on the search space and force the agent to `mimic' the expert \cite{abbeel2004apprenticeship,abbeel2008apprenticeship}, which does not meet our expectation.

\subsection{Mirror Stimuli Function}

To address this problem, we design a \textbf{mirror stimuli function}, denoting as $F_{topology}$, which aims to $softly$ guide the agent while preventing a hard and strong constraint on the search space. The design of the mirror stimuli function is inspired by the \emph{mirror neuron system} in primate's premotor cortex. This system is responsible for the linkage of self-generated and observed demonstrations. The mirror neuron fires both when an animal acts and when the animal observes the same action performed by another, which is an important scheme for learning new skills by imitation. In our problem, the mirror stimuli function has a similar functionality as the mirror neuron. Given the architecture sampled by the agent as the self-generated demonstration, the expert network as the observed demonstration, our mirror stimuli function will output a signal to judge the topological similarity between these two networks. The higher output represents higher similarity, where the highest for the exact expert network.

The mirror stimuli function is defined as a linear function of feature count:
\begin{eqnarray}\label{eq:mirror-s-f}
F_{topology}(m) =  w^T \cdot \mu ,
\end{eqnarray}
where $w \in \mathbb{R}^{k\times 1}$. Such a linear parametric form is easy to optimize, while effective enough to use as the evaluation of topology, as further shown in our experiment.

By substituting Equation \ref{eq:mu} to Equation \ref{eq:mirror-s-f}, we can get
\begin{eqnarray}
F_{topology}(m) = \sum_{t=1}^{T} \gamma^{t} \cdot w^T \cdot \phi(s_t).
\end{eqnarray}
Thus, the problem of solving the parameter $w$ could be regarded as the problem of finding a time-step reward function $r(s_t) = w^T \cdot \phi(s_t)$, whose corresponding policy has a maximum value at the sequence of expert network (i.e., the value of $F_{topology}(m^*)$, $m^*= \{ {(s_t^*, a_t^*)}\}_{t=1...T}$ denotes the expert network). This refers to the standard inverse reinforcement learning problem.

To find such an reward function, we use the feature match algorithm proposed in \cite{abbeel2004apprenticeship}. For the expert network, the architecture is generated following an expert policy $\pi^*$, which has a maximum value for the following expression:
\begin{equation}
\begin{aligned}
J_{\pi^*} &= \mathbb E_{\pi^*}[\sum_{t=1}^{T} \gamma^{t} r(s_t)] = w^T \cdot \mathbb E_{\pi^*}[\sum_{t=1}^{T} \gamma^{t} \phi(s_t)]  \\
&= w^T \cdot \mathbb E_{\pi^*}[\mu] = w^T \cdot M_{\pi^*}.
\end{aligned}
\end{equation}
As we have one expert network, $M_{\pi^*}$ is estimated as $M_{\pi^*} = \mathbb E_{\pi^*}[\mu] \approx \mu^* = \sum_{t=1}^{T} \gamma^{t} \phi(s_t^*)$.

To get the weight parameter $w$ of the unknown reward function $r(s_t)$, we need to find a policy $\hat{\pi}$ whose performance is close to that of the expert's:
\begin{eqnarray}
\begin{aligned}
|J_{\hat{\pi}} - J_{\pi^*}| &= |w^T \cdot M_{\hat{\pi}} - w^T \cdot M_{\pi^*}| \le \epsilon.
\end{aligned}
\end{eqnarray}

This process could be regarded as `imitating' the observed behavior in the mirror neuron system, which makes the self-generated demonstration (regarded as $J_{\hat{\pi}}$) similar to the observed demonstration (regarded as $J_{\pi^*}$). So the problem is reduced to finding a policy $\hat{\pi}$ that induces the expectation of feature count $M_{\hat{\pi}}$ close to $M_{\pi^*}$. This feature matching problem could be solved by max-margin optimization, derived as,
\begin{eqnarray}
\max_{w:\|w\|_2 \le 1} \min_{\forall \hat{\mu}} {  w^T \cdot M_{\pi^*} - w^T \cdot M_{\hat{\pi}}}.
\end{eqnarray}
Thus the weight parameter $w$ is optimized following:
\begin{eqnarray}
\begin{aligned} \label{eq:optimize}
\max_{\delta,w}  &\qquad \delta \\
s.t. & \qquad w^T \cdot M_{\pi^*} \ge {w}^T \cdot M_{\hat{\pi}} + \delta, \quad \forall \ \hat{\pi} \\
 & \qquad \|w\|_2 \le 1.
\end{aligned}
\end{eqnarray}
The detailed algorithm is illustrated in Algorithm \ref{alg:1}.

\begin{algorithm}[tb]
\caption{\small{Max-Margin Optimization for Inverse Reinforcement Learning}}
\label{alg:1}
\begin{algorithmic}
\small{
\STATE {set $i=1$, randomly pick policy $\hat{\pi}_0$, compute $\hat{M}_0$;} 
\REPEAT 
\STATE Compute $\delta^{(i)}$ in optimization problem of Equation \ref{eq:optimize} with $\{ \hat{M} \} = \{ \hat{M}_j, j=0...i-1 \}$, get $w^{(i)}$, $\delta^{(i)}$;
\STATE Using standard RL algorithm, find the optimal policy as $\hat{\pi}_i$ with reward function $r^{(i)}(s) = (w^{(i)})^T \cdot \phi(s)$;
\STATE Compute $\hat{M}_i$;
\STATE $i = i + 1$;
\UNTIL{$\delta^{(i)} \le \epsilon $} 
\STATE return $w$;}
\end{algorithmic}
\end{algorithm}

During the agent's training stage, we add the output of mirror stimuli function as an additional reward term. The complete reward function in Section \ref{section-formulation} is calculated as:
\begin{eqnarray}\label{eq:function}
R(m) = F_{accuracy}(m) + \lambda F_{topology}(m),
\end{eqnarray}
where $F_{accuracy}(m)$ denotes model $m$'s accuracy percentage on target task, $\lambda$ denotes a balance scalar. 

By optimizing this multi-objective search problem, the agent is guided by both the topological similarity and the accuracy. Thus, the agent can efficiently explore the search space to generate high-speed, topologically elegant architectures along with high accuracy.

\subsection{Search Space and Training Strategy}\label{section-search}
In this section we introduce the search space and training strategy of our IRLAS. We will further discuss the generalization of our mirror stimuli function to other typical architecture search approaches in Section \ref{section:generalize}. In our IRLAS, the search space consists of operations based on their prevalence in the CNN literature. The considered operations are: \emph{Depthwise convolution} with kernel size 1$\times$1, 3$\times$3, 5$\times$5; \emph{Max pooling} with kernel size 3$\times$3, 5$\times$5; \emph{Average pooling} with kernel size 3$\times$3, 5$\times$5; \emph{Identity}; \emph{Elemental add} with two input layers; and \emph{Concat} with two input layers. Note that the depthwise convolution operation refers to pre-activation convolution containing ReLU, convolution and batch normalization. All the layers without successor in the searched block are concatenated together as the final output.

For the searching stage, we utilize Q-learning method to train the agent to take actions that maximize the cumulative reward, which is formulated as Equation \ref{eq:function}. 
Q-learning iteratively updates the action-selection policy following the Bellman Equation:
\begin{eqnarray}\label{eq:Bellman}
Q({s_t}, {a_t}) = r_t + \gamma\max_{a^{'}}Q({s_{t+1}}, {a^{'}}),
\end{eqnarray}
where $r_t$ denotes the intermediate reward observed for the current state $s_t$. Since $r_t$ could not be explicitly measured, reward shaping method is used, derived as $r_t = R(m) / T$, where $T$ denotes the state length referring to the number of layers. The Bellman Equation is achieved following Temporal-Difference control algorithm:
\begin{eqnarray}\label{eq:Bellman-update}
\begin{aligned}
Q({s_t}, {a_t}) = & (1-\eta)Q(s_t ,a_t )\\
 &+\eta[r_{t+1} + \gamma\max_{a^{'}}Q({s_{t+1}}, {a^{'}})],
\end{aligned}
\end{eqnarray}
where $\eta$ denotes the learning rate.

The whole learning procedure is summarized as follows: The agent first samples a network architecture, which is taken as input of the mirror stimuli function. Then the generated network is trained on a certain task to get the validation accuracy. The reward, which is the combination of the accuracy and the output value of the mirror stimuli function, is used to update the Q-value. The above process circulates for iterations and the agent learns to sample block structure with higher accuracy and more elegant topology iteratively.

\subsection{Generalization of Mirror Stimuli Function}
\label{section:generalize}

It is worthy to point out that our mirror stimuli function can be easily generalized to different architecture search algorithms. For algorithms that involve architecture sampling and performance evaluation for the sampled architecture, including reinforcement learning based methods and evolutionary methods, we can simply utilize the output of Equation \ref{eq:function} as an alternative of evaluation, while the other searching steps remain the same to the original algorithm. The only difference lies in the expression of state feature function $\phi(s)$, which need to be modified due to different candidate operations in the search space of different algorithms. Thus, the topological information is considered during the searching process. 

For differentiable architecture search algorithm, typically DARTS \cite{liu2018darts}, the architecture is encoded by a set of continuous variables $\alpha=\{\alpha^{\{i,j\}}\}$ ($(i,j)$ denotes a pair of nodes, i.e. a path in the architecture). Thus, the weight parameters and architecture parameters could be trained jointly via standard gradient descent. To introduce topological information to the training procedure in differentiable architecture search algorithms, we add an additional loss term $L_{topology}$ calculated by mirror stimuli function to the original cross entropy loss. To convert the continuous $\alpha$ to discreted architectures, we consider the $softmax$ output of $\alpha$ as a probabilistic distribution of all possible architectures, denoted as $\{p_k\}$, and sample according to the distribution to get state feature $\phi(s)$. Since the conversion from architecture parameters $\alpha$ to state feature $\phi(s)$ is non-differentiable, the output of mirror stimuli function cannot be backpropagated. Here, we consider the solution based on REINFORCE algorithm \cite{williams1992simple}, so the loss term $L_{topology}$ is calculated and updated as:
\begin{eqnarray}
\begin{aligned}
L_{topology} &= \sum_{k=1}^{K}p_k F_{topology}(m_k) \\
\nabla L_{topology} &\approx \frac{1}{K}\sum_{k=1}^{K}F_{topology}(m_k)\nabla log(p_k),
\end{aligned}
\end{eqnarray}
where $K$ is the number of sampled architectures.

\section{Experiments and Results}
\subsection{Implementation Details}

In this section, we introduce the implementation details of our IRLAS. We use a distributed asynchronous framework as proposed in \cite{zhong2018practical}, which enables efficient network generation on multiple machines with multiple GPUs. With this framework, our IRLAS can sample and train networks in parallel to speed up the whole training process. For the inverse reinforcement learning procedure, ResNet, whose convolution operation is modified to depthwise convolution, is chosen as the expert network to calculate the weight $w$ in the mirror stimuli function. The training procedure is about 3 hours on CPU.

For our IRLAS, we choose Q-value table as the agent. We use Q-learning with epsilon-greedy and experience replay buffer. 
At each training iteration, the agent samples 64 structures with their corresponding rewards from the memory to update Q-values following Equation \ref{eq:Bellman-update}. For the hyperparameters of Q-learning process, the learning rate $\eta$ is set to 0.01, the discount factor $\gamma$ is 0.9 and the balance scalar $\lambda$ is 30. The mini-batch size is set to 64 and the maximum layer index for a block is set to 24. The agent is trained for 180 iterations, which totally samples 11,500 blocks. Each sampled architecture is trained with fixed 12 epochs with Adam optimizer to get evaluation of $F_{accuracy}$.

We also generalize our mirror stimuli function to the different architecture search algorithm. We choose DARTS \cite{liu2018darts} as the basic algorithm. The additional loss term $L_{topology}$ is scaled by 0.5 and added to the original cross-entropy loss. The number of the sampled architectures $K$ is set to 5. All the other training details and hyperparameters follow the original paper. For both of the conditions, the architecture searching processes are proposed on dataset CIFAR-10 \cite{krizhevsky2009learning}. 

\subsection{Results}

\paragraph{Results on CIFAR-10}
\label{section:CIFAR}

\begin{table}[tb]
\centering

\caption{\small{IRLAS's results compared with state-of-the-art methods on CIFAR-10 dataset. ``Error'' is the top-1 misclassification rate on the CIFAR-10 test set, ``Param'' is the number of model parameters.}}
\label{tb:cifar}

\small
\begin{tabular}{c | c | c }
    \hline
    {Method}  & Param & Error(\%)\\
    \hline
    \hline
    Resnet \cite{he2016deep}                              &  1.7M  & 6.61 \\
    Resnet (pre-activation) \cite{he2016identity}         & 10.2M  & 4.62 \\
    Wide ResNet \cite{zagoruyko2016wide}                  & 36.5M  & 4.17 \\
    DenseNet (k=12) \cite{huang2017densely}               &  1.0M  & 5.24 \\
    DenseNet (k=12) \cite{huang2017densely}               &  7.0M  & 4.10 \\
    DenseNet (k=24) \cite{huang2017densely}               & 27.2M  & 3.74 \\
    DenseNet-BC (k=40) \cite{huang2017densely}            & 25.6M  & 3.46 \\
    \hline
    \hline
    MetaQNN (top model) \cite{baker2016designing}         & 11.2M  & 6.92 \\
    NAS v1 \cite{zoph2016neural}                          &  4.2M  & 5.50 \\
    EAS \cite{cai2018efficient}                           & 23.4M  & 4.23 \\
    Block-QNN-A, N=4 \cite{zhong2018practical}            &     -  & 3.60 \\
    Block-QNN-S, N=2 \cite{zhong2018practical}            &  6.1M  & 3.30 \\
    NASNet-A (6 @ 768) \cite{zoph2017learning}            &  3.3M  & 2.65 \\
    NASNet-B (4 @ 1152) \cite{zoph2017learning}           &  2.6M  & 3.73 \\
    NASNet-C (4 @ 640) \cite{zoph2017learning}            &  3.1M  & 3.59 \\
    PNASNet-5 \cite{liu2017progressive}                   &  3.2M  & 3.41 \\
    ENAS \cite{pham2018efficient}                         &  4.6M  & 2.89 \\
    AmoebaNet-A \cite{real2018regularized}                &  3.2M  & 3.34 \\
    DARTS \cite{liu2018darts}                             &  3.4M  & 2.83 \\
    \hline
    \hline
    IRLAS                                                 & 3.91M  & 2.60 \\
    IRLAS-differential                                    & 3.43M  & 2.71 \\
    \hline 
\end{tabular}
 \vspace{-14pt}
\end{table}

After the searching process, we select the searched optimal block structure and train the network on CIFAR-10 until convergence. In this phase, the training data is augmented with randomly cropping size of $32\times32$, horizontal flipping and Cutout \cite{devries2017improved}. The cosine learning rate scheme is utilized with the initial learning rate of 0.2. The momentum rate is set to 0.9 and weight decay is set to 0.0005. All the networks are trained for 600 epochs with 256 batch size.

For the task of image classification on CIFAR-10, we set the total number of stacked blocks as 10. The results are reported in Table \ref{tb:cifar} along with other models. We see that our proposed IRLAS achieves a 2.60\% test error, which shows a state-of-the-art performance over both human-designed networks and auto-generated networks. For the differential setting, the result is reported in Table \ref{tb:cifar} as IRLAS-differential. Compared to the result reported in original paper (2.83\% error rate), the searched architecture facilitated by our mirror stimuli function achieves a higher accuracy.

\paragraph{Results on ImageNet}\label{section-imagenet}

For the ImageNet task, we transfer the model searched on CIFAR-10 by increasing the total number of stacked blocks and the filter channel size. We consider the \emph{mobile} setting to compare inference speed. The training is conducted with a mini-batch size of 256 with input image size $224\times224$. Randomly cropping and flipping are used to augment data. We choose SGD strategy for optimization with cosine learning rate scheme. The accuracy on test images is evaluated with center crop. 
We use the true inference latency for fair comparison, which is validated for 16 batch size on TensorRT \cite{tensorRT} framework with one Titan Xp. 

The results are illustrated in Table \ref{tb:imagenet-mobile}. Our IRLAS-mobile achieves a state-of-the-art accuracy over both the human-designed and auto-generated architectures. As for the inference latency, our IRLAS-mobile can achieve 2$\sim$4$\times$ fewer inference latency compared with most auto-generated architectures benefiting from the elegant topology facilitated by our mirror stimuli function. We also further squeeze the number of stacked blocks of IRLAS-mobile and increase conduct a IRLAS-mobile-fast model with an inference speed of 9ms, making our model even faster than human-designed network MobileNetV2. Note that MnasNet \cite{tan2018mnasnet} was searched directly on ImageNet dataset and need to validate time latency during searching, which is a very resource-exhausted process due to the high training cost on such a large scale dataset. As the shuffle operation, channel split operation and inverted block backbone used in ShuffleNetV2 and MobileNet-224 are not adopted in our search space, we believe our inference speed can be further boosted by introducing them to our searching process.

\begin{table}[tb]
\centering
\caption{\small{ImageNet classification results in the \emph{mobile} setting. The input image size is 224$\times$224. The inference latency is validated with 16 batch size on TensorRT framework.}}
\label{tb:imagenet-mobile}
\small
\begin{tabular}{c | c | c   }
    \hline
    {Method} & Latency  & Acc (\%) \\
    \hline
    Inception V1 \cite{szegedy2015going}                     &   -   & 69.8 \\
    MobileNet-224 \cite{howard2017mobilenets}                &   6ms & 70.6 \\
    ShuffleNet \cite{hluchyj1991shuffle}                     &  10ms & 70.9 \\
    MobileNetV2 1.4 \cite{sandler2018mobilenetv2}            &  10ms & 74.7 \\
    ShuffleNetV2 2$\times$ \cite{ma2018shufflenet}          &   6ms  & 74.9 \\
    \hline
    \hline
    NASNet-A(4 @ 1056) \cite{zoph2017learning}               &  23ms & 74.0 \\
    AmoebaNet-A \cite{real2018regularized}                   &  33ms & 74.5 \\
    PNASNet \cite{liu2017progressive}                        &  25ms & 74.2 \\
    DARTS \cite{liu2018darts}                                &  55ms & 73.1 \\
    MnasNet \cite{tan2018mnasnet}                            &  11ms & 74.79 \\
    \hline
    \hline
    IRLAS-mobile                                             &  12ms & 75.28\\
    IRLAS-mobile-fast                                        &   9ms & 75.15\\
    \hline 
\end{tabular}
\vspace{-14pt}
\end{table}

\subsection{Analysis of Inverse Reinforcement Learning}
 
In this section, we conduct an analysis of inverse reinforcement learning algorithm. As we introduce inverse reinforcement learning to avoid the agent to exhibit preference on the expert network, we compare the output value changes of our mirror stimuli function with those of the feature count $\mu$ by modifying a specific architecture. Here we choose the expert architecture ResNet, and modify it in three ways: $Modify 1$, adding a $conv 3\times3$ operation before the residual function; $Modify 2$, adding a $conv 3\times3$ operation after the residual function; $Modify 3$, removing the short-cut connection. The results are illustrated in Figure \ref{fig-weight} (a). Since $Modify 1$ and $Modify 2$ have a minor change in topology than $Modify 3$, our mirror stimuli function is able to output relative value change, where the feature count is very sensitive to tiny changes. As a result, comparing to direct feature count, our mirror stimuli function is a more reasonable guidance to avoid the agent to just mimic the expert network, which helps the agent to explore the search space without being overly restricted.

\begin{figure*}
  \begin{minipage}{1.0\linewidth}
  \centering{{}
      \includegraphics[width=0.90\linewidth]{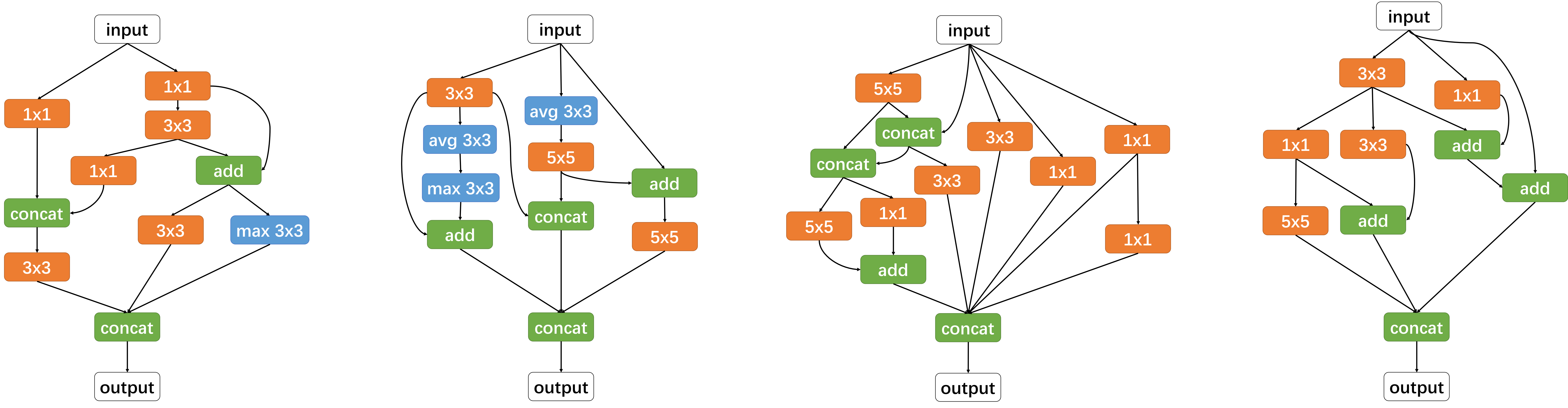}}
  \caption{\small{Topologies of top-4 block architectures searched \emph{without} mirror stimuli function.}}
  \label{fig-compare_1}
  \end{minipage} \hfill
  \begin{minipage}{1.0\linewidth}
    \centering{
      \includegraphics[width=0.9\linewidth]{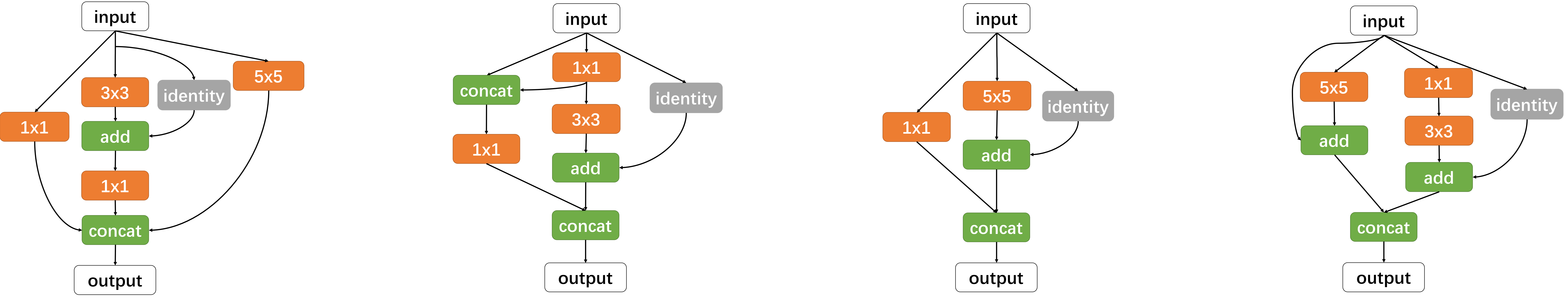}}
  \caption{\small{Topologies of top-4 block architectures searched \emph{with} mirror stimuli function.}}
  \label{fig-compare_2}
  \end{minipage} \hfill

  \vspace{-14pt}
\end{figure*}

\subsection{Search Efficiency}

\begin{figure}[tb]
  \begin{center}
    \includegraphics[width=1.0\linewidth]{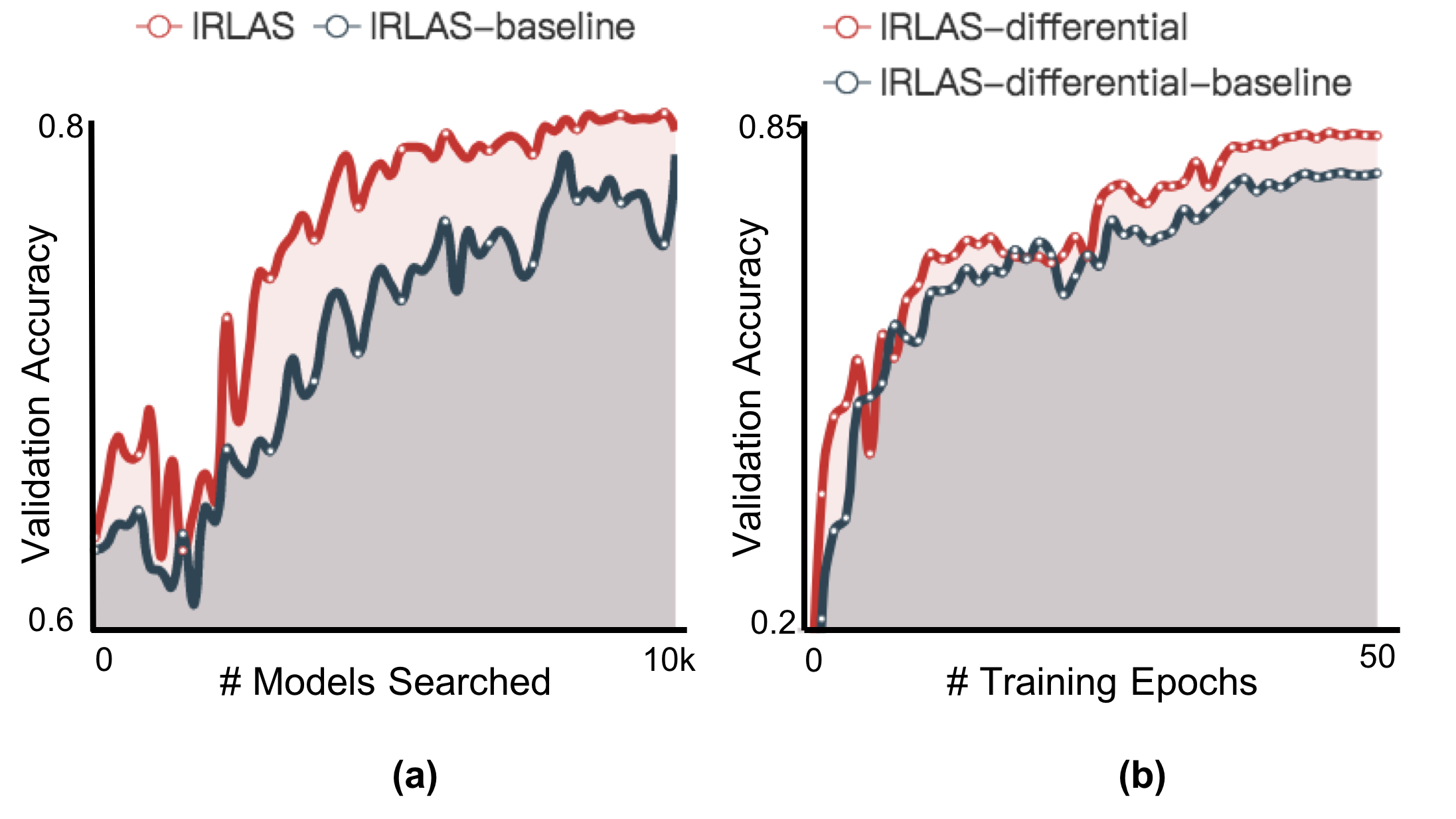}
  \end{center}
  \vspace{-14pt}
  \caption{\small{Convergence curves for the searching processes comparing with searching without mirror stimuli function. For both of the conditions, our methods converge faster benefiting from the guidance provided by the expert network's topology.}}
  \label{fig-converge}
  \vspace{-14pt}
\end{figure}

In this section, we perform an analysis on search efficiency. Note that the overall searching cost largely depends on the design of search strategy, which is orthogonal to the design of our mirror stimuli function. 
To illustrate the efficiency improvement introduced by our mirror stimuli function, we conducted two experiments based on two search algorithms of different kinds: one is BlockQNN \cite{zhong2018practical}, the other is DARTS \cite{liu2018darts}. For each experiment, the baseline followed the searching process proposed in original paper, compared with the searching facilitated by our mirror stimuli function. We evaluate the efficiency of by mirror stimuli function by comparing the \emph{relative} improvement of convergence speed, instead of the \emph{absolute} search time. Convergence curves are reported in Figure \ref{fig-converge}. For both of the conditions, our methods converge faster, benefiting from the guidance provided by the expert network's topology. The results further demonstrate that our mirror stimuli function is able to be generalized to different search algorithms and improve the search efficiency.

\subsection{Ablation Study}

In this section, we perform analysis to illustrate how mirror stimuli function affects the topology of final searched architecture. We first illustrate topologies of top-4 block architectures searched without and with mirror stimuli function in Figure \ref{fig-compare_1} and Figure \ref{fig-compare_2}. It is obvious that architectures searched without mirror stimuli function are complicated, including numerous operations and connections, while our searched models are much more simple and elegant. Furthermore, our searched models are more topologically similar to ResNet, each containing a shortcut following add operation to form the residual function.

\begin{figure}[tb]
  \begin{center}
    \includegraphics[width=1.0\linewidth]{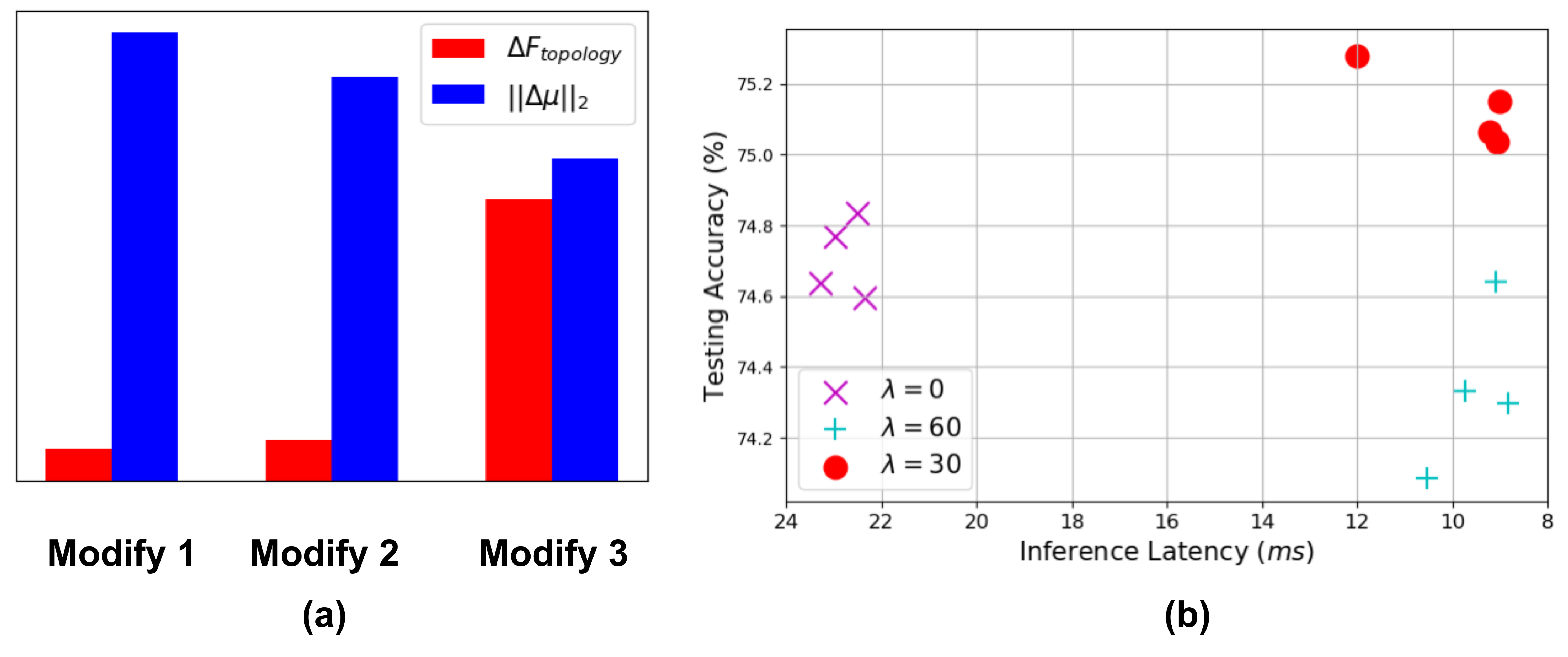}
  \end{center}
  \caption{\small{(a) Comparison the output value changes of mirror stimuli function and feature count for three modified models. (b) Results of inference latency and accuracy on ImageNet of 4 top models from each experiment with different $\lambda$. $\lambda=30$ is used in our IRLAS to balance the trade-off between accuracy and speed.}}
  \label{fig-weight}
  \vspace{-14pt}
\end{figure}

We further conduct IRLAS with three different $\lambda$: 0, 30, 60. All three searching experiments followed the same procedure described in Section \ref{section-search}. For each experiment, top-4 models were chosen and transfered to meet the ImageNet mobile setting, with about 5M parameters. These models were then trained from scratch on ImageNet, following settings in Section \ref{section-imagenet}. The final inference latency and accuracy of these models are illustrated in Figure \ref{fig-weight} (b). It can be noticed that the inference speed of searched architectures can be drastically improved by utilizing mirror stimuli function, about 1$\times$ faster. For $\lambda=60$, the prior topological knowledge of expert network is too strong for searching, which results in accuracy drop. $\lambda=30$ is regarded as a choice to balance the trade-off between accuracy and speed.

\section{Conclusion}
In this paper, we have proposed an inverse reinforcement learning method for architecture search. Based on the knowledge learned from the expert-designed networks, our mirror stimuli function can provide topological guidance to architecture search, which can be easily generalized to different architecture search algorithms. Inverse reinforcement learning method has been introduced to train this function, helping the agent to efficiently explore the large search space without being overly restricted. Experiment results have shown that our proposed IRLAS achieves to search high-speed architectures with high accuracy. How to extract representation of multiple networks to further improve the performance seems to be an interesting future work. 

{\small
\bibliographystyle{ieee}
\bibliography{egbib}

\begin{thebibliography}{10}\itemsep=-1pt

\bibitem{tensorRT}
https://developer.nvidia.com/tensorrt.

\bibitem{abbeel2008apprenticeship}
P.~Abbeel, D.~Dolgov, A.~Y. Ng, and S.~Thrun.
\newblock Apprenticeship learning for motion planning with application to
  parking lot navigation.
\newblock In {\em Intelligent Robots and Systems, 2008. IROS 2008. IEEE/RSJ
  International Conference on}, pages 1083--1090. IEEE, 2008.

\bibitem{abbeel2004apprenticeship}
P.~Abbeel and A.~Y. Ng.
\newblock Apprenticeship learning via inverse reinforcement learning.
\newblock In {\em ICML}, page~1. ACM, 2004.

\bibitem{baker2016designing}
B.~Baker, O.~Gupta, N.~Naik, and R.~Raskar.
\newblock Designing neural network architectures using reinforcement learning.
\newblock {\em arXiv preprint arXiv:1611.02167}, 2016.

\bibitem{brock2017smash}
A.~Brock, T.~Lim, J.~M. Ritchie, and N.~Weston.
\newblock Smash: one-shot model architecture search through hypernetworks.
\newblock {\em arXiv preprint arXiv:1708.05344}, 2017.

\bibitem{cai2018efficient}
H.~Cai, T.~Chen, W.~Zhang, Y.~Yu, and J.~Wang.
\newblock Efficient architecture search by network transformation.
\newblock AAAI, 2018.

\bibitem{choi2013bayesian}
J.~Choi and K.-E. Kim.
\newblock Bayesian nonparametric feature construction for inverse reinforcement
  learning.
\newblock In {\em IJCAI}, pages 1287--1293, 2013.

\bibitem{deng2009imagenet}
J.~Deng, W.~Dong, R.~Socher, L.-J. Li, K.~Li, and L.~Fei-Fei.
\newblock Imagenet: A large-scale hierarchical image database.
\newblock In {\em CVPR}, pages 248--255, 2009.

\bibitem{devries2017improved}
T.~DeVries and G.~W. Taylor.
\newblock Improved regularization of convolutional neural networks with cutout.
\newblock {\em arXiv preprint arXiv:1708.04552}, 2017.

\bibitem{domhan2015speeding}
T.~Domhan, J.~T. Springenberg, and F.~Hutter.
\newblock Speeding up automatic hyperparameter optimization of deep neural
  networks by extrapolation of learning curves.
\newblock In {\em IJCAI}, volume~15, pages 3460--8, 2015.

\bibitem{he2016deep}
K.~He, X.~Zhang, S.~Ren, and J.~Sun.
\newblock Deep residual learning for image recognition.
\newblock In {\em CVPR}, pages 770--778, 2016.

\bibitem{he2016identity}
K.~He, X.~Zhang, S.~Ren, and J.~Sun.
\newblock Identity mappings in deep residual networks.
\newblock In {\em ECCV}, pages 630--645. Springer, 2016.

\bibitem{hluchyj1991shuffle}
M.~G. Hluchyj and M.~J. Karol.
\newblock Shuffle net: An application of generalized perfect shuffles to
  multihop lightwave networks.
\newblock {\em Journal of Lightwave Technology}, 9(10):1386--1397, 1991.

\bibitem{howard2017mobilenets}
A.~G. Howard, M.~Zhu, B.~Chen, D.~Kalenichenko, W.~Wang, T.~Weyand,
  M.~Andreetto, and H.~Adam.
\newblock Mobilenets: Efficient convolutional neural networks for mobile vision
  applications.
\newblock {\em arXiv preprint arXiv:1704.04861}, 2017.

\bibitem{huang2017densely}
G.~Huang, Z.~Liu, L.~Van Der~Maaten, and K.~Q. Weinberger.
\newblock Densely connected convolutional networks.
\newblock In {\em CVPR}, volume~1, page~3, 2017.

\bibitem{krizhevsky2009learning}
A.~Krizhevsky and G.~Hinton.
\newblock Learning multiple layers of features from tiny images.
\newblock Technical report, Citeseer, 2009.

\bibitem{krizhevsky2014cifar}
A.~Krizhevsky, V.~Nair, and G.~Hinton.
\newblock The cifar-10 dataset.
\newblock {\em online: http://www. cs. toronto. edu/kriz/cifar. html}, 2014.

\bibitem{levine2011nonlinear}
S.~Levine, Z.~Popovic, and V.~Koltun.
\newblock Nonlinear inverse reinforcement learning with gaussian processes.
\newblock In {\em NIPS}, pages 19--27, 2011.

\bibitem{liu2017progressive}
C.~Liu, B.~Zoph, M.~Neumann, J.~Shlens, W.~Hua, L.-J. Li, L.~Fei-Fei,
  A.~Yuille, J.~Huang, and K.~Murphy.
\newblock Progressive neural architecture search.
\newblock In {\em ECCV}, September 2018.

\bibitem{liu2018darts}
H.~Liu, K.~Simonyan, and Y.~Yang.
\newblock Darts: Differentiable architecture search.
\newblock {\em arXiv preprint arXiv:1806.09055}, 2018.

\bibitem{ma2018shufflenet}
N.~Ma, X.~Zhang, H.-T. Zheng, and J.~Sun.
\newblock Shufflenet v2: Practical guidelines for efficient cnn architecture
  design.
\newblock {\em arXiv preprint arXiv:1807.11164}, 2018.

\bibitem{perez2018efficient}
J.-M. Perez-Rua, M.~Baccouche, and S.~Pateux.
\newblock Efficient progressive neural architecture search.
\newblock {\em arXiv preprint arXiv:1808.00391}, 2018.

\bibitem{pham2018efficient}
H.~Pham, M.~Y. Guan, B.~Zoph, Q.~V. Le, and J.~Dean.
\newblock Efficient neural architecture search via parameter sharing.
\newblock {\em arXiv preprint arXiv:1802.03268}, 2018.

\bibitem{ratliff2006maximum}
N.~D. Ratliff, J.~A. Bagnell, and M.~A. Zinkevich.
\newblock Maximum margin planning.
\newblock In {\em ICML}, pages 729--736. ACM, 2006.

\bibitem{real2018regularized}
E.~Real, A.~Aggarwal, Y.~Huang, and Q.~V. Le.
\newblock Regularized evolution for image classifier architecture search.
\newblock {\em arXiv preprint arXiv:1802.01548}, 2018.

\bibitem{sandler2018mobilenetv2}
M.~Sandler, A.~Howard, M.~Zhu, A.~Zhmoginov, and L.-C. Chen.
\newblock Mobilenetv2: Inverted residuals and linear bottlenecks.
\newblock In {\em CVPR}, pages 4510--4520, 2018.

\bibitem{saxena2016convolutional}
S.~Saxena and J.~Verbeek.
\newblock Convolutional neural fabrics.
\newblock In {\em NIPS}, pages 4053--4061, 2016.

\bibitem{szegedy2015going}
C.~Szegedy, W.~Liu, Y.~Jia, P.~Sermanet, S.~Reed, D.~Anguelov, D.~Erhan,
  V.~Vanhoucke, and A.~Rabinovich.
\newblock Going deeper with convolutions.
\newblock In {\em CVPR}, pages 1--9, 2015.

\bibitem{szegedy2016rethinking}
C.~Szegedy, V.~Vanhoucke, S.~Ioffe, J.~Shlens, and Z.~Wojna.
\newblock Rethinking the inception architecture for computer vision.
\newblock In {\em CVPR}, pages 2818--2826, 2016.

\bibitem{tan2018mnasnet}
M.~Tan, B.~Chen, R.~Pang, V.~Vasudevan, and Q.~V. Le.
\newblock Mnasnet: Platform-aware neural architecture search for mobile.
\newblock {\em arXiv preprint arXiv:1807.11626}, 2018.

\bibitem{williams1992simple}
R.~J. Williams.
\newblock Simple statistical gradient-following algorithms for connectionist
  reinforcement learning.
\newblock {\em Machine learning}, 8(3-4):229--256, 1992.

\bibitem{xie2017genetic}
L.~Xie and A.~L. Yuille.
\newblock Genetic cnn.
\newblock In {\em ICCV}, pages 1388--1397, 2017.

\bibitem{xie2017aggregated}
S.~Xie, R.~Girshick, P.~Doll{\'a}r, Z.~Tu, and K.~He.
\newblock Aggregated residual transformations for deep neural networks.
\newblock In {\em CVPR}, pages 5987--5995. IEEE, 2017.

\bibitem{zagoruyko2016wide}
S.~Zagoruyko and N.~Komodakis.
\newblock Wide residual networks.
\newblock {\em arXiv preprint arXiv:1605.07146}, 2016.

\bibitem{zhong2018practical}
Z.~Zhong, Z.~Yang, B.~Deng, J.~Yan, W.~Wu, J.~Shao, and C.-L. Liu.
\newblock Blockqnn: Efficient block-wise neural network architecture
  generation.
\newblock {\em arXiv preprint arXiv:1808.05584}, 2018.

\bibitem{ziebart2008maximum}
B.~D. Ziebart, A.~L. Maas, J.~A. Bagnell, and A.~K. Dey.
\newblock Maximum entropy inverse reinforcement learning.
\newblock In {\em AAAI}, volume~8, pages 1433--1438. Chicago, IL, USA, 2008.

\bibitem{zoph2016neural}
B.~Zoph and Q.~V. Le.
\newblock Neural architecture search with reinforcement learning.
\newblock {\em arXiv preprint arXiv:1611.01578}, 2016.

\bibitem{zoph2017learning}
B.~Zoph, V.~Vasudevan, J.~Shlens, and Q.~V. Le.
\newblock Learning transferable architectures for scalable image recognition.
\newblock {\em arXiv preprint arXiv:1707.07012}, 2(6), 2017.

\end{thebibliography}
}

\end{document}